# Google's Cloud Vision API Is Not Robust To Noise

Hossein Hosseini, Baicen Xiao and Radha Poovendran

Network Security Lab (NSL), Department of Electrical Engineering, University of Washington, Seattle, WA
{hosseinh, bcxiao, rp3}@uw.edu

*Abstract*—Google has recently introduced the Cloud Vision API for image analysis. According to the demonstration website, the API "quickly classifies images into thousands of categories, detects individual objects and faces within images, and finds and reads printed words contained within images." It can be also used to "detect different types of inappropriate content from adult to violent content."

In this paper, we evaluate the robustness of Google Cloud Vision API to input perturbation. In particular, we show that by adding sufficient noise to the image, the API generates completely different outputs for the noisy image, while a human observer would perceive its original content. We show that the attack is consistently successful, by performing extensive experiments on different image types, including natural images, images containing faces and images with texts. For instance, using images from ImageNet dataset, we found that adding an average of $14.25\%$ impulse noise is enough to deceive the API. Our findings indicate the vulnerability of the API in adversarial environments. For example, an adversary can bypass an image filtering system by adding noise to inappropriate images. We then show that when a noise filter is applied on input images, the API generates mostly the same outputs for restored images as for original images. This observation suggests that cloud vision API can readily benefit from noise filtering, without the need for updating image analysis algorithms.

## I. INTRODUCTION

In recent years, Machine Learning (ML) techniques have been extensively deployed for computer vision tasks, particularly visual classification problems, where new algorithms reported to achieve or even surpass the human performance [1]–[3]. Success of ML algorithms has led to an explosion in demand. To further broaden and simplify the use of ML algorithms, cloud-based services offered by Amazon, Google, Microsoft, BigML, and others have developed ML-as-a-service tools. Thus, users and companies can readily benefit from ML applications without having to train or host their own models.

Recently, Google introduced the Cloud Vision API for image analysis [4]. A demonstration website has been also launched, where for any selected image, the API outputs the image labels, identifies and reads the texts contained in the image and detects the faces within the image. It also determines how likely is that the image contains inappropriate contents, including adult, spoof, medical, or violence contents.

The implicit assumption in designing and developing ML models is that they will be deployed in noise-free and benign settings. Real-world sensors, however, suffer from noise, blur

This work was supported by ONR grants N00014-14-1-0029 and N00014-16-1-2710, ARO grant W911NF-16-1-0485 and NSF grant CNS-1446866.

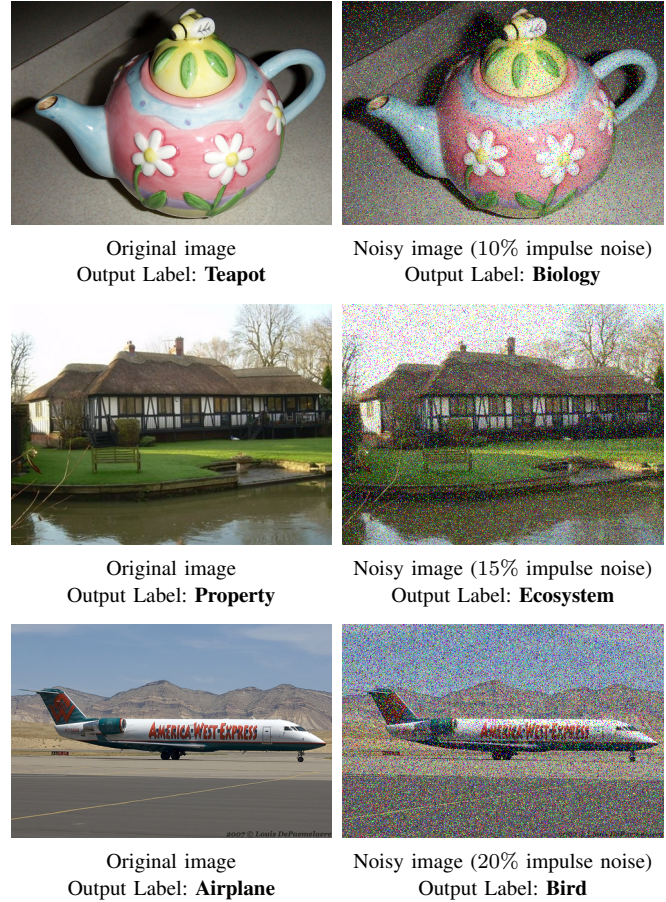

Fig. 1: Illustration of the attack on Google Cloud Vision API. By adding sufficient noise to the image, we can force the API to output completely different labels. Captions are the labels with the highest confidence returned by the API. For noisy images, none of the output labels are related to corresponding original images. Images are chosen from the ImageNet dataset.

and other imperfections. Hence, designing computer vision models to be robust is imperative for real-world applications, such as banking, medical diagnosis, and autonomous driving. Moreover, recent research have pointed out the vulnerability of ML models in adversarial environments [5]–[7]. Security evaluation of ML systems is an emerging field of study. Several papers have presented attacks on various ML systems, such as voice interfaces [8], face-recognition systems [9], toxic comment detectors [10], and video annotation systems [11].

In this paper, we evaluate the robustness of Google Cloud Vision API to input perturbations. [1] In particular, we investigate whether we can modify an image in such a way that a human observer would perceive its original content, but the API generates different outputs for it. For modifying the images, we add either impulse noise or Gaussian noise to them. Due to the inherent low-pass filtering characteristic of the humans vision system, humans are capable of perceiving image contents from images slightly corrupted by noise [12].

Our experimental results show that by adding sufficient noise to the image, the API is deceived into returning labels which are not related to the original image. Figure 1 illustrates the attack by showing original and noisy images along with the most confident labels returned by the API. We show that the attack is consistently successful, by performing extensive experiments on different image types, including natural images, images containing faces and images with texts. Our findings indicate the vulnerability of Google cloud vision API in real-world applications. For example, a driveless car may wrongly identify the objects in rainy weather. Moreover, the API can be subject to attacks in adversarial environments. For example, a search engine may suggest irrelevant images to users, or an image filtering system can be bypassed by adding noise to inappropriate images.

We then evaluate different methods for improving the robustness of the API. Since we only have a black-box access to the API, we assess whether noise filtering can improve the API performance on noisy inputs, while maintaining the accuracy on clean images. Our experimental results show that when a noise filter is applied on input images, the API generates mostly the same outputs for restored images as for original images. This observation suggests that the cloud vision API can readily benefit from noise filtering, without the need for updating the image analysis algorithms.

The rest of this paper is organized as follows. Section II reviews related literature and Section III presents noise models. The proposed attack on Google cloud vision API is given in Section IV. Section V describes some countermeasures to the attack and Section VI concludes the paper.

## II. Related Work

Several papers have recently showed that the performance of deep convolutional neural networks drops when the model is tested on distorted inputs, such as noisy or blurred images [13]–[15]. For improving the robustness of machine learning models to input perturbations, an end-to-end architecture is proposed in [16] for joint denoising, deblurring, and classification. In [17], the authors presented a training method to stabilize deep networks against small input distortions. It has been also observed that augmenting training data with perturbed images can enhance the model robustness [13], [18]. In contrast, in this paper we demonstrate the vulnerability of a real-world image classifier system to input perturbations.

[1]The experiments are performed on the interface of Google Cloud Vision API's website on Apr. 7, 2017.

We also show that the model robustness can be improved by applying a noise filter on input images, thus without the need for fine-tuning the model.

The noisy images used in our attack can be viewed as a form of adversarial examples [19]. An adversarial example is defined as a modified input, which causes the classifier to output a different label, while a human observer would recognize its original content. Note that we could deceive the could vision API without having any knowledge about the learning algorithm. Also, unlike the existing black-box attacks on learning systems [20], [21], we have no information about the training data or even the set of output labels of the model. Moreover, unlike the current methods for generating adversarial examples [22], we perturb the input completely *randomly*, which results in a more serious attack vector in real-world applications.

## III. Image Noise

A color image $x$ is a three-dimensional array of pixels $x_{i,j,k}$, where $(i,j)$ is the image coordinate and $k \in \{1,2,3\}$ denotes the coordinate in color space. In this paper, we encode the images in RGB color space. Most image file formats use 24 bits per pixel (8 bits per color channel), which results in 256 different colors for each color space. Therefore, the minimum and maximum values of each pixel are 0 and 255, respectively, which correspond to the darkest and brightest colors.

For modifying the images, we add either impulse noise or Gaussian noise to them. These noise types often occur during image acquisition and transmission [23]. Impulse Noise, also known as Salt-and-Pepper Noise, is commonly modeled by [24]:

$$\tilde{x}_{i,j,k} = \begin{cases} 0 & \text{with probability } \frac{p}{2} \\ x_{i,j,k} & \text{with probability } 1-p \\ 255 & \text{with probability } \frac{p}{2} \end{cases}$$

where $x$, $\tilde{x}$ and $p$ are the original and noisy images and the *noise density*, respectively. Impulse noise can be removed using spatial filters which exploit the correlation of adjacent pixels. We use the weighted-average filtering method, proposed in [24], for restoring images corrupted by impulse noise.

A noisy image corrupted by Gaussian noise is obtained as $\hat{x}_{i,j,k} = x_{i,j,k} + z$, where $z$ is a zero-mean Gaussian random variable. The pixel values of the noisy image should be clipped, so that they remain in the range of 0 to 255. Gaussian noise can be reduced by filtering the input with low-pass kernels [23].

For assessing the quality of the restored image $x^*$ compared to original image $x$, we use the Peak Signal-to-Noise Ratio (PSNR). For images of size $d_1 \times d_2 \times 3$, PSNR value is computed as follows [25]:

$$PSNR = 10 \cdot \log_{10} \left( \frac{255^2}{\frac{1}{3 d_1 d_2} \sum_{i,j,k}(x_{i,j,k} - x^*_{i,j,k})^2} \right).$$

PSNR value is measured in dB. Typical values for the PSNR are usually considered to be between 20 and 40 dB, where higher is better [26].

## IV. THE PROPOSED ATTACK ON CLOUD VISION API

In this section, we describe the attack on Google Cloud Vision API. The goal of the attack is to modify a given image in such a way that the API returns completely different outputs than the ones for original image, while a human observer would perceive its original content. We perform the experiments on different image types, including natural images from the ImageNet dataset [27], images containing faces from the Faces94 dataset [28], and images with text. When selecting an image for analysis, the API outputs the image labels, detects the faces within the image, and identifies and reads the texts contained in the image.

The attack procedure is as follows. We first test the API with the original image and record the outputs. We then test the API with a modified image, generated by adding very low-density impulse noise. If we can force the API to output completely different labels, or to fail to detect faces or identify the texts within the image, we declare the noisy image as the adversary's image. Otherwise, we increase the noise density and retry the attack. We continue to increase the noise density until we can successfully force the API to output wrong labels. In experiments, we start the attack with 5% impulse noise and increase the noise density each time by 5%.

Figure 1 shows the API's output label with the highest confidence score, for the original and noisy images. As can be seen, unlike the original images, the API wrongly labels the noisy images, despite that the objects in noisy images are easily recognizable. Trying on 100 images of the ImageNet dataset, we needed on average 14.25% impulse noise density to deceive the cloud vision API. Figure 2 shows the adversary's success rate versus the noise density. As can be seen, by adding 35% impulse noise, the attack always succeeded on the samples from ImageNet dataset.

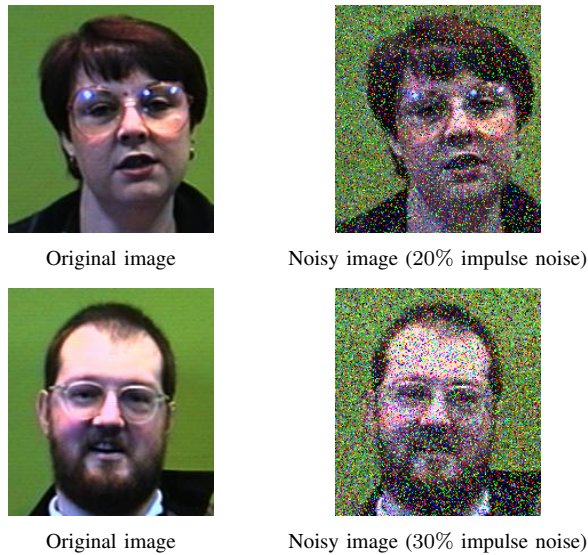

Fig. 3: Images of faces, chosen from the Faces94 dataset, and their noisy versions. Unlike the original images, cloud vision API fails to detect the face in noisy images.

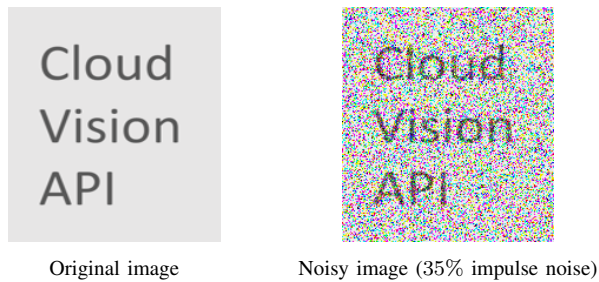

Fig. 4: An images with text and its noisy version. Unlike the original image, cloud vision API fails to identify any texts in noisy image.

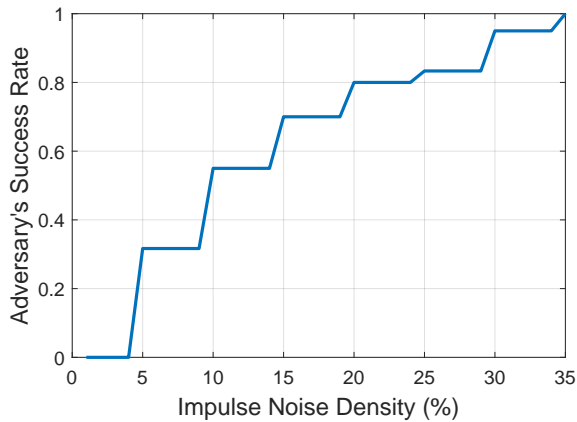

Fig. 2: Adversary's success rate versus the impulse noise density for sample images from ImageNet dataset. By adding 35% impulse noise, the attack always succeeds in changing the API's output labels.

Figure 3 shows sample images from the Faces94 dataset and the corresponding noisy images. Unlike the original images, the API fails to detect the face in noisy ones. Trying on the first 20 images of each female and male categories, we needed on average 23.8% impulse noise density to deceive the cloud vision API. Similarly, figure 4 shows an image with text and the corresponding noisy image. The API correctly reads the text within the original image, but fails to identify any texts in the noisy one, despite that the text within the noisy image is easily readable.

We also tested the API with images corrupted by Gaussian noise and obtained similar results as impulse noise. That is, by adding zero-mean Gaussian noise with sufficient variance, we can always force the API to generate a different output than the one for the original image, while a human observer would perceive its original content.

## V. COUNTERMEASURES

The success of our attack indicates the importance of designing the learning system to be robust to input perturbations. It has been shown that the robustness of ML algorithms can be improved by using regularization or data augmentation during training [29]. In [30], the authors proposed adversarial training,

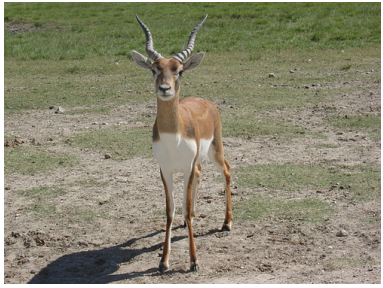
(a) Original image

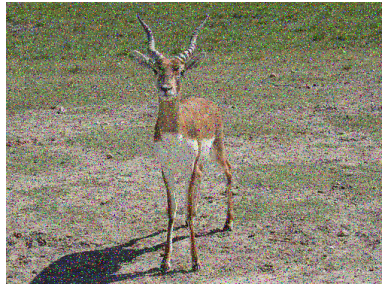
(b) Noisy image (10% impulse noise)

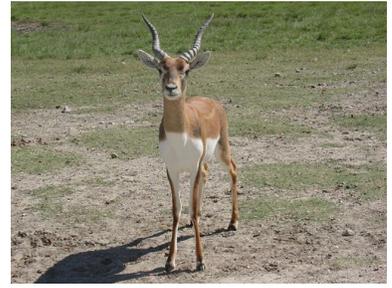
(c) Restored image (PSNR = 33 dB)

(d) API's output labels for original image.

(e) API's output labels for noisy image.

(f) API's output labels for restored image.

Fig. 5: Screenshots of the labels returned by cloud vision API for original, noisy and restored images. The original image is chosen from ImageNet dataset. None of the labels returned for the noisy image are related to labels of the original image, while labels of the restored image are mostly the same as the ones for original image.

which iteratively creates a supply of adversarial examples and includes them into the training data. Approaches based on robust optimization however may not be practical, since the model needs to be retrained.

For image recognition algorithms, a more viable approach is preprocessing the inputs. Natural images have special properties, such as high correlation among adjacent pixels, sparsity in transform domain or having low energy in high frequencies [23]. Noisy inputs typically do not lie in the same space as natural images. Therefore, by projecting the input image down to the space of natural images, which is often done by passing the image through a filter, we can reverse the effect of the noise or adversarial perturbation.

We assess the performance of the cloud vision API when a noise filter is applied before the image analysis algorithms. We did the experiments on all the sample images from ImageNet and Faces94 datasets, corrupted by either impulse or Gaussian noise. Restored images are generated by applying the weighted-average filter [24] for impulse noise and a low-pass filter for Gaussian noise. In all cases, when testing on the restored image, the API generates mostly the same outputs as for the original image.

Figure 5 shows the screenshots of the API's output labels for original, noisy and restored images of a sample image from ImageNet dataset. As can be seen, none of the labels returned for the noisy image are related to labels of the original image. However, the labels of the restored image are mostly the same as the ones for original image.

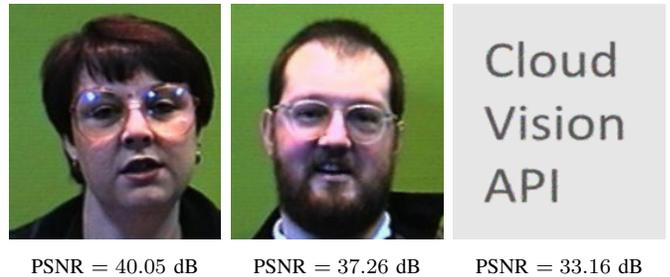

PSNR = 40.05 dB    PSNR = 37.26 dB    PSNR = 33.16 dB

Fig. 6: The restored images, generated by applying the weighted-average filter [24] on the noisy images of figures 3 and 4. Captions show the PSNR values with respect to the original images. Although the API fails to detect the face in the noisy face images, it correctly detects the same face attributes for restored images as the original images. Also, unlike the noisy version of the text image, the API correctly reads the text within the restored image.

Similarly, figure 6 shows restored images of the images with faces from figure 3 and the image with text from figure 4. Unlike the noisy images, the API correctly detects the same face attributes for restored face images as original images, and can read the text within the restored text image. The results suggest that the cloud vision API can readily benefit from noise filtering prior to applying image analysis algorithms.

## VI. CONCLUSION

In this paper, we showed that Google Cloud Vision API can be easily deceived by an adversary without compromising the system or having any knowledge about the specific details of the algorithms used. In essence, we found that by adding noise, we can always force the API to output irrelevant labels or to fail to detect any face or text within the image. We also showed that when testing with the restored images, the API generates mostly the same outputs as for the original images. This suggests that the system's robustness can be readily improved by applying a noise filter on the inputs, without the need for updating the image analysis algorithms.

## REFERENCES

[1] A. Krizhevsky, I. Sutskever, and G. E. Hinton, "Imagenet classification with deep convolutional neural networks," in *Advances in neural information processing systems*, pp. 1097–1105, 2012.
[2] K. Simonyan and A. Zisserman, "Very deep convolutional networks for large-scale image recognition," *arXiv preprint arXiv:1409.1556*, 2014.
[3] C.-Y. Lee, S. Xie, P. W. Gallagher, Z. Zhang, and Z. Tu, "Deeply-supervised nets.," in *AISTATS*, vol. 2, p. 5, 2015.
[4] https://cloud.google.com/vision/.
[5] L. Huang, A. D. Joseph, B. Nelson, B. I. Rubinstein, and J. Tygar, "Adversarial machine learning," in *Proceedings of the 4th ACM workshop on Security and artificial intelligence*, pp. 43–58, ACM, 2011.
[6] N. Papernot, P. McDaniel, S. Jha, M. Fredrikson, Z. B. Celik, and A. Swami, "The limitations of deep learning in adversarial settings," in *Security and Privacy (EuroS&P), 2016 IEEE European Symposium on*, pp. 372–387, IEEE, 2016.
[7] D. Amodei, C. Olah, J. Steinhardt, P. Christiano, J. Schulman, and D. Mané, "Concrete problems in ai safety," *arXiv preprint arXiv:1606.06565*, 2016.
[8] N. Carlini, P. Mishra, T. Vaidya, Y. Zhang, M. Sherr, C. Shields, D. Wagner, and W. Zhou, "Hidden voice commands," in *25th USENIX Security Symposium (USENIX Security 16), Austin, TX*, 2016.
[9] M. Sharif, S. Bhagavatula, L. Bauer, and M. K. Reiter, "Accessorize to a crime: Real and stealthy attacks on state-of-the-art face recognition," in *Proceedings of the 2016 ACM SIGSAC Conference on Computer and Communications Security*, pp. 1528–1540, ACM, 2016.
[10] H. Hosseini, S. Kannan, B. Zhang, and R. Poovendran, "Deceiving google's perspective api built for detecting toxic comments," *arXiv preprint arXiv:1702.08138*, 2017.
[11] H. Hosseini, B. Xiao, and R. Poovendran, "Deceiving google's cloud video intelligence api built for summarizing videos," *arXiv preprint arXiv:1703.09793*, 2017.
[12] F. Röhrbein, P. Goddard, M. Schneider, G. James, and K. Guo, "How does image noise affect actual and predicted human gaze allocation in assessing image quality?," *Vision research*, vol. 112, pp. 11–25, 2015.
[13] I. Vasiljevic, A. Chakrabarti, and G. Shakhnarovich, "Examining the impact of blur on recognition by convolutional networks," *arXiv preprint arXiv:1611.05760*, 2016.
[14] S. Karahan, M. K. Yildirum, K. Kirtac, F. S. Rende, G. Butun, and H. K. Ekenel, "How image degradations affect deep cnn-based face recognition?," in *Biometrics Special Interest Group (BIOSIG), 2016 International Conference of the*, pp. 1–5, IEEE, 2016.
[15] S. Dodge and L. Karam, "Understanding how image quality affects deep neural networks," in *Quality of Multimedia Experience (QoMEX), 2016 Eighth International Conference on*, pp. 1–6, IEEE, 2016.
[16] S. Diamond, V. Sitzmann, S. Boyd, G. Wetzstein, and F. Heide, "Dirty pixels: Optimizing image classification architectures for raw sensor data," *arXiv preprint arXiv:1701.06487*, 2017.
[17] S. Zheng, Y. Song, T. Leung, and I. Goodfellow, "Improving the robustness of deep neural networks via stability training," in *Proceedings of the IEEE Conference on Computer Vision and Pattern Recognition*, pp. 4480–4488, 2016.
[18] S. Dodge and L. Karam, "Quality resilient deep neural networks," *arXiv preprint arXiv:1703.08119*, 2017.
[19] C. Szegedy, W. Zaremba, I. Sutskever, J. Bruna, D. Erhan, I. Goodfellow, and R. Fergus, "Intriguing properties of neural networks," *arXiv preprint arXiv:1312.6199*, 2013.
[20] N. Papernot, P. McDaniel, I. Goodfellow, S. Jha, Z. B. Celik, and A. Swami, "Practical black-box attacks against deep learning systems using adversarial examples," *arXiv preprint arXiv:1602.02697*, 2016.
[21] H. Hosseini, Y. Chen, S. Kannan, B. Zhang, and R. Poovendran, "Blocking transferability of adversarial examples in black-box learning systems," *arXiv preprint arXiv:1703.04318*, 2017.
[22] N. Carlini and D. Wagner, "Towards evaluating the robustness of neural networks," *arXiv preprint arXiv:1608.04644*, 2016.
[23] A. C. Bovik, *Handbook of image and video processing*. Academic press, 2010.
[24] H. Hosseini, F. Hessar, and F. Marvasti, "Real-time impulse noise suppression from images using an efficient weighted-average filtering," *IEEE Signal Processing Letters*, vol. 22, no. 8, pp. 1050–1054, 2015.
[25] R. H. Chan, C.-W. Ho, and M. Nikolova, "Salt-and-pepper noise removal by median-type noise detectors and detail-preserving regularization," *IEEE Transactions on image processing*, vol. 14, no. 10, pp. 1479–1485, 2005.
[26] A. Amer, A. Mitiche, and E. Dubois, "Reliable and fast structure-oriented video noise estimation," in *Image Processing. 2002. Proceedings. 2002 International Conference on*, vol. 1, pp. I–I, IEEE, 2002.
[27] J. Deng, W. Dong, R. Socher, L.-J. Li, K. Li, and L. Fei-Fei, "Imagenet: A large-scale hierarchical image database," in *Computer Vision and Pattern Recognition, 2009. CVPR 2009. IEEE Conference on*, pp. 248–255, IEEE, 2009.
[28] "Face recognition data, university of essex, uk, face 94." http://cswww.essex.ac.uk/mv/allfaces/faces94.html.
[29] U. Shaham, Y. Yamada, and S. Negahban, "Understanding adversarial training: Increasing local stability of neural nets through robust optimization," *arXiv preprint arXiv:1511.05432*, 2015.
[30] I. J. Goodfellow, J. Shlens, and C. Szegedy, "Explaining and harnessing adversarial examples," *arXiv preprint arXiv:1412.6572*, 2014.